\documentclass{article}
\usepackage{eufrak}
\usepackage{enumitem}
\setitemize{noitemsep,topsep=0pt,parsep=0pt,partopsep=0pt}
\usepackage{times}

\makeatletter
\newcommand\footnoteref[1]{\protected@xdef\@thefnmark{\ref{#1}}\@footnotemark}
\makeatother

\usepackage{microtype}
\usepackage{graphicx}
\usepackage{subfigure}
\usepackage{booktabs} %
\usepackage{multirow}

\usepackage{todonotes}

\usepackage{amsmath}
\usepackage{amsfonts}
\usepackage{amsthm}
\usepackage{siunitx}

\newcommand{\real}{\mathbb{R}}

\newtheoremstyle{icml}
  {\topsep}
  {\topsep}
  {\normalfont}
  {0pt}
  {\bfseries}
  {\quad}
  { }
  {\thmname{#1}\thmnumber{ #2}\thmnote{ \normalfont(#3)}}
\theoremstyle{icml}
\newtheorem{property}{Property}
\newtheorem*{property*}{Property}
\newtheorem{lemma}{Lemma}
\newtheorem*{lemma*}{Lemma}

\newtheorem*{example*}{Example}

\newtheorem*{remark*}{Remark}

\usepackage{multirow}
\usepackage{mathtools}
\usepackage{array, caption}
\usepackage{graphicx}
\usepackage{makecell}
\usepackage{arydshln}

\usepackage{dsfont}

\usepackage{natbib}

\usepackage{algorithm}
\usepackage{algorithmic}

\usepackage{xcolor}
\usepackage{amsmath}

\newcommand{\brown}[1]{\textcolor{black}{#1}}
\newcommand{\mat}[1]{\mathbf{\uppercase{#1}}}
\renewcommand{\vec}[1]{\mathbf{#1}}

\newcommand{\highlight}[1]{(\textbf{#1})}

\definecolor{SkyBlue}{RGB}{0,0,0}
\newcommand{\SkyBlue}[1]{\emph{\textcolor{SkyBlue}{#1}}}

\definecolor{tbcolor}{RGB}{0,0,0}
\newcommand{\tbcolor}[1]{\textcolor{tbcolor}{#1}}

\newcommand{\para}[1]{\fontsize{10.245}{12}\selectfont\textbf{#1\rule{0.7em}{0ex}}}

\usepackage{hyperref}

\usepackage[accepted]{icml2019}

\icmltitlerunning{Weakly-Supervised Temporal \textit{Lo}calization via Occurrence \textit{Co}unt Learning}

\usepackage{dblfloatfix}
\begin{document} 

\twocolumn[
\icmltitle{Weakly-Supervised Temporal \textit{Lo}calization via Occurrence \textit{Co}unt Learning}

\begin{icmlauthorlist}
\icmlauthor{Julien Schroeter}{ed}
\icmlauthor{Kirill Sidorov}{ed}
\icmlauthor{David Marshall}{ed}
\end{icmlauthorlist}

\icmlaffiliation{ed}{Cardiff University, United Kingdom}

\icmlcorrespondingauthor{Julien Schroeter}{SchroeterJ1@cardiff.ac.uk}

\icmlkeywords{Machine learning, ICML, Weakly Supervised, Localization, Deep Learning, Music Transcription, Occurrence Count}

\vskip 0.3in
]



\printAffiliationsAndNotice{}  

\begin{abstract} 
We propose a novel model for temporal detection and localization which allows the training of deep neural networks using only counts of event occurrences as training labels. 
This powerful weakly-supervised framework alleviates the burden of the imprecise and time-consuming process of annotating event locations in temporal data.
Unlike existing methods, in which localization is explicitly achieved by design, our model learns localization implicitly as a byproduct of learning to count instances.
This unique feature is a direct consequence of the model's theoretical properties.
We validate the effectiveness of our approach in a number of experiments (drum hit and piano onset detection in audio, digit detection in images) and demonstrate performance comparable to that of fully-supervised state-of-the-art methods, despite much weaker training requirements.

\end{abstract}

\section{Introduction}
\label{intro}
In recent years, deep learning techniques have demonstrated outstanding performance on numerous tasks ranging from object recognition to natural language processing~\cite{schmidhuber2015deep, lecun2015deep}. However, this success comes at a cost:~large annotated datasets are typically needed for training. Alleviating this requirement remains an important open problem. Indeed, while hand-labeling can be a very tedious and time-consuming process~\cite{imagenet_cvpr09}, automated label assignments based on external sources~\cite{abu2016youtube} are not always available nor reliable. Both approaches suffer further from the same inherent risk of introducing errors and imprecisions into datasets~\cite{frenay2014classification}. 

In order to address this issue, some approaches attempt to extract a signal from unlabeled data hence making the training effectively unsupervised. While unsupervised models have been successfully leveraged for representation learning \cite{lee2009convolutional, doersch2015unsupervised, radford2015unsupervised}, dimensionality reduction \cite{hinton2006reducing} or clustering of highly structured data \cite{hastie2009unsupervised}, the absence of any annotation limits their versatility and effectiveness on more complex tasks.

Another angle of attack consists in weakening the annotation requirement. Indeed, weakly-supervised models are an effective way to mitigate the challenge of accurate and cost-effective dataset labeling by being able to leverage simplified and more easily accessible labels. Annotations describing only instance classes, but not positions (in space or time), have been successfully leveraged in computer vision for object detection \cite{fergus2003object} or action localization \cite{duchenne2009automatic}. In recent years, a number of deep learning approaches have pushed the state-of-the-art further \cite{bilen2016weakly, shou2018autoloc}. However, despite its potential, the weakly-supervised framework has only recently attracted attention in other domains, such as audio event localization~\cite{kumar2016audio}.

In this paper, we propose a novel weakly-supervised learning approach for localization and detection of events in sequential data, which requires \emph{only the number of event occurrences} as training label. Unlike its fully-supervised counterparts, our model successfully learns both event representation and detection without any localization prior.
\begin{figure}[t!]
\begin{center}
\centerline{\includegraphics[width=8.3cm]{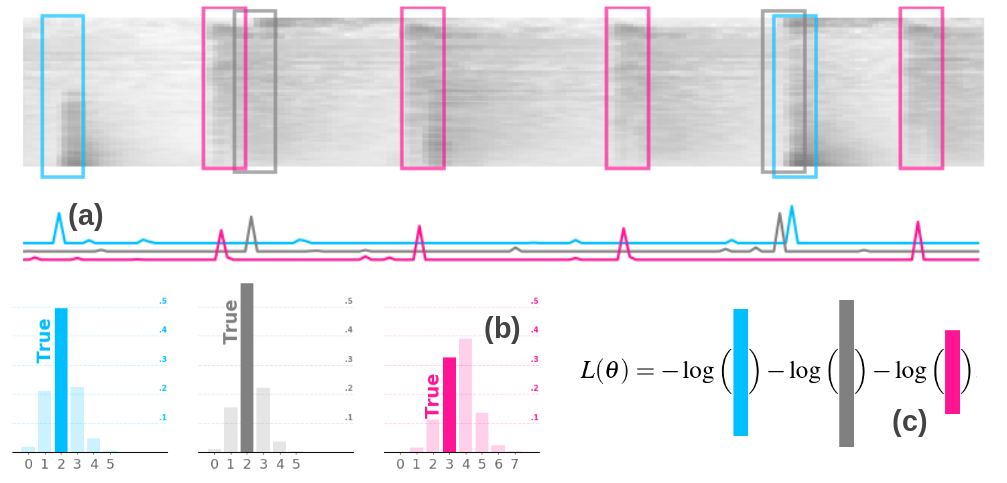}}
\caption{Illustration of loss computation for drum transcription. Given as input a spectrogram, the model successively estimates: (a) Event processes $\vec{p}_{i}$, (b) Count distributions $Y_{i}$, (c) Loss $L(\theta)$.}
\label{HeaderImage}
\end{center}
\vskip -0.30in
\end{figure} 

The need for temporally localizing instantaneous events using only occurrence counts for training arises in many different contexts. For example, in music, the number of note instances can easily be counted, whereas determining the exact onset time requires far more effort. Similarly, in sports, the number of occurrences of numerous types of events (e.g., goals, fouls, \textit{etc.}) is commonly known in aggregated forms at the end of games, while precise localization of these actions is substantially more tedious.

Unlike previous works which are explicitly designed for localization, our proposed model learns event localization implicitly via model constraints. More precisely, our model does not rely on any attention mechanisms~\cite{mnih2014recurrent, luong2015effective} or other devices for explicit localization, but rather \textit{indirectly} learns localization as a byproduct of learning to count instances (Section~\ref{sec:sectionModel}). After some implementation considerations in Section~\ref{sec:sectionImplementation}, the model is thoroughly evaluated on several tasks such as drum transcription, piano note onset detection, and digit detection in Sections~\ref{DrumTrascription}, \ref{sectionPiano}, and \ref{sectionDigit} respectively.

\para{Contributions}
In this work, we:~(a) Propose a novel model to solve weakly-supervised temporal localization and detection tasks.
(b) Present an analysis of the model's properties. 
(c) Demonstrate the efficiency of our weakly-supervised model on several experiments: drum transcription and piano note onset detection.
(d) Develop an extension of our model to object detection in images.

\section{Related Work} 
The majority of weakly-supervised temporal localization models have been developed for both video and audio events localization. We briefly review them below.

\para{Weakly-supervised video action localization}
First, Duchenne et al.~\yrcite{duchenne2009automatic} proposed a discriminative clustering approach to segment action snippets from the background. This clustering framework was later revisited by Bojanowski et al.~\yrcite{bojanowski2014weakly} to handle temporal assignment problems~--- i.e., to partition the sequence using an ordered list of actions. This problem was also addressed by Huang et al.~\yrcite{huang2016connectionist} using an extended Connectionist Temporal Classification method~\cite{graves2006connectionist} and by Richard et al.~\yrcite{richard2017weakly} introducing a fine-grained subaction model. 

Another prevailing problem in action localization consists in action intervals prediction rather than temporal segmentation. Initial works include the unsupervised generative bag of spatio-temporal interest points approach proposed by Niebles~et~al.~\yrcite{niebles2008unsupervised} which actually infers a more general spatio-temporal localization. The same problem was tackled by Nguyen~et~al.~\yrcite{nguyen2009weakly} by simultaneously learning segment selection and classification. Later, Gan~et~al.~\yrcite{gan2015devnet} used spatio-temporal saliency maps obtained by back-passing through the classification CNN to achieve localization, while Singh~and~Lee~\yrcite{singh2017hide} extended their Hide-and-Seek approach to action localization. Recently, attention-based approaches have been used extensively. First, Wang~et~al.~\yrcite{wang2017untrimmednets} introduced the UntrimmedNet~--- an attention model performing localization on pre-selected video segments. The mechanism was improved by Nguyen~et~al.~\yrcite{nguyen2017weakly} with class-specific activation maps, while Shou et~al.~\yrcite{shou2018autoloc} replaced the fixed thresholding by a dynamic approach based on the proposed Outer-Inner-Contrastive (OIC) loss. 

However, by focusing on subsegments regardless of their temporality, most methods neglect additional temporal information contained in the data (e.g., relative order of events, what precedes an event, \textit{etc.}). To address this issue, Niebles et~al.~\yrcite{niebles2010modeling} modeled actions as a composition of motion segments. In this paper, as the core of our approach relies on recurrent units, the temporal nature of the data is intrinsically taken into account.

\vskip 0.08in

\para{Weakly-supervised audio localization}
Similarly to action localization, attention-based models have become a common solution to weakly-supervised audio localization tasks. Xu et al. improved their own attention-based convolutional recurrent neural network \cite{xu2017attention} by applying a trainable gated linear unit instead of the classical ReLU~\cite{xu2018large}, while Kong et al.~\yrcite{kong2017joint} performed joint detection and classification on overlapping blocks. Alternatively, Kumar~and~Raj~\yrcite{kumar2016audio} leveraged multiple instance learning to address the localization task. A similar method based on convolutional network rather than support vector machines or classical neural networks was proposed later by Liu~and~Yang~\yrcite{liu2016event}. Lee~et~al.~\yrcite{lee2017ensemble} further improved the framework by incorporating segment-level and clip-level predictions ensembling. Finally, Adavanne~and~Virtanen~\yrcite{adavanne2017sound} used a stacked convolutional and recurrent network to sequentially predict stronger and weaker labels.

Overall, as for actions in videos, weakly-supervised localization in audio is also commonly achieved explicitly using attention mechanisms or segment-level detection and classification. This contrasts with our approach which \emph{implicitly learns localization} while the occurrence count is being learned. Another unique feature of our model resides in the temporal nature of the events: this paper focuses on localizing instantaneous events precisely (sometimes in the order of milliseconds) rather than estimating the extent of longer actions.

\section{Problem Formulation}
\label{sectionProblem}

Since the task at hand is slightly different to that of most works on temporal event localization, we begin by formally defining the data assumptions and the exact localization problem.

Let $\mathcal{D}$ be the training data with $N$ samples:
\begin{equation}
\mathcal{D} := \left\lbrace\left(\mat{X}_{i},\vec{y}_{i}\right): 0<i\leq N\right\rbrace.
\end{equation}
Let us consider the relationship between predictor $\mat{X}_{i}$ and dependent variables $\vec{y}_{i}$. First, each $\mat{x}_{i}$ is assumed to be an observable temporal sequence, i.e., $\mat{x}_{i} = \left(\vec{x}_{i}\!\left(t\right)\right)_{t=1}^{T_{i}} \in {\rm I\!R}^{T_{i} \times \lambda}$. (Depending on the application, this can stand for any $\lambda$-dimensional time-series such as spectrograms or DNN-learned representations.) Second, we assume there exists an underlying unobservable event process $ \mat{e}_{i} = \left(\vec{e}_{i}\!\left(t\right)\right)_{t=1}^{T_{i}} \in \left\lbrace 0,1 \right\rbrace ^{T_{i}\times d}$, indicating the presence of events. (For ease of explanation, instead of general multivariate event processes we consider the univariate case ($d=1$) throughout the theoretical part of this work.) Each event process is assumed to be a function of its predictors:
\begin{equation}
\big(\vec{e}_{i}\left(t\right)\big)_{t=1}^{\tau}=g\Big( \big(\vec{x}_{i}\!\left(t\right)\big)_{t=1}^{\tau} \Big), \forall \tau\le T_i.
\end{equation}
 Finally, the observable dependent variables $\vec{y}_{i}$ are defined as the total number of occurred events:
\begin{equation}
\vec{y}_{i} = \sum_{t} \vec{e}_{i}\!\left( t \right).
\end{equation}
Informally, the main objective of this paper consists in accurately detecting and localizing instantaneous events (whose duration is one time step) in time, while only requiring the total number of occurrences as training data. Hence, the problem we want to solve is the following:

\textbf{Event process estimation (EPE) problem:} \textit{Estimate the event process $\mat{e} = \left(\vec{e}\left(t\right)\right)_{t=1}^{T_{j}} \in \left\lbrace 0,1 \right\rbrace ^{T}$ underlying an unseen process $\mathbf{X}$ using only the data $\mathcal{D}$ for training.}

\begin{remark*}
In addition to localization in unseen data, the capability of the model to estimate event processes from counts is itself useful, for example for enriching the training data.
\end{remark*}

\subsection{Underlying Assumptions}

The definition above implies only weak assumptions about the event processes:

{\bf Uniqueness} \emph{Only one event per class occurs at each time step.} This condition is commonly met in most datasets. Otherwise, the use of smaller temporal granularities can easily solve the problem.

{\bf Localization} \emph{Each event lasts for a unique time step.} This assumption sets our approach apart from previous works on temporal localization. In this setting, events are by definition not spread out in time. Nevertheless, this localization assumption only needs to hold in the representation space: $\vec{x}_{i}(t)$ may correspond to representations of longer time-intervals in the original data space.

\section{Model}
\label{sec:sectionModel}
The main idea of this paper is to design a model such that localization implicitly emerges by constraint:  the model is intrinsically bound to output a clear-cut estimate of an event process in order to make a valid prediction of the number of occurrences.

\subsection{Model Definition}
\label{sectionDefinition}

We propose the following model to solve the
event process estimation (EPE) problem:
\begin{equation}
\begin{split}
Y_{i} & =\sum_{t} E_{i}\!\left( t \right),
\\
E_{i} \left( t \right) & = \mathfrak{B} \left( p_{i}\!\left( t \right) \right) \text{, ind. Bernoulli},
\\
p_{i}\left( t \right) &= f \left( \big(\vec{x}_{i}\!\left(n\right)\big)_{n=1}^{t} \right).
\end{split}
\end{equation}
The event occurrences $e_{i}$ and counts $y_i$ are realizations of $E_{i}$ and the (stochastic) count distributions $Y_i$ respectively. (The independence assumption of the Bernoulli distributions is valid even though the probabilities $p_i$ might be correlated.)
In this paper, the function $f$ will be estimated using recurrent units, such as an LSTM~\cite{hochreiter1997long} or GRU~\cite{cho2014learning} with model parameters $\theta$, which constitutes a rather natural choice given the model's temporal structure:
\begin{equation}
\hat{p}_{i,\theta} \left( t \right)  = \hat{f}_{\theta} \left( \big(\vec{x}_{i}\!\left(n\right)\big)_{n=1}^{t} \right).
\end{equation}

\subsection{Loss}
\label{Loss}
By definition, the event count distribution $Y_{i,\theta}$ follows a Poisson-binomial distribution:
\begin{equation}
\Pr(Y_{i,\theta}=k\mid\mat{x}_{i}) = \sum\limits_{A\in F_k} \prod\limits_{l \in A} \hat{p}_{i, \theta}\!\left( l \right) \prod\limits_{j\in A^c} (1-\hat{p}_{i,\theta}\! \left( j \right) ),
\label{poissonBinomialDistribution}
\end{equation}
where $F_{k}$ is the set of all subsets of $\lbrace 1,2,...,T_{i} \rbrace$ of size $k$.

Thus, estimation of the parameter set $\theta$ can be done by comparing the distribution $\Pr(Y_{i,\theta}\!=\!k\!\mid\! \mat{x}_{i})$ to the target sample distribution determined by~$y_i$. The Kullback-Leibler divergence~\yrcite{kullback1951information}, which in this specific case corresponds to the cross-entropy and max-likelihood, is a suitable choice for the loss function:
\begin{equation}
L(\theta) = -\sum_{i}  \log  \left( \Pr \left( Y_{i,\theta}= y_{i}\mid \mat{x}_{i}  \right)\right).
\end{equation}
The \brown{computation process of this newly introduced LoCo-loss (\textbf{Lo}calization through \textbf{Co}unting)} is illustrated in \mbox{Figure~\ref{HeaderImage}}.

\subsection{Properties}

The convergence of the proposed model towards a stable solution of the EPE problem is not obvious. Below, we show how optimizing $L(\theta)$ leads not only to an exact estimation of the total number of occurrences but also to a precise estimation of the event processes.

\begin{property}[Mass shift irreversibility]~\\
\vskip 0.00in
\centerline{$\big({Y}_{i,\theta}\!\left(t\right)\big)_{t=1}^{T_{i}}$ is monotonically increasing.}
\end{property}

This statement directly follows from the definition of $Y$ as a sum of non-negative random variables. Intuitively, this implies that any probability mass shift towards increasing count values can never be shifted back --- constituting a strong implicit model constraint. Thus, unlike most of its counterparts, our model is prevented from early triggering since all mass movements are \brown{irreversible. Indeed, if the model was anticipating, the loss would surge whenever events would ultimately not occur, as counts estimates cannot be reduced \textsl{a posteriori}. \highlight{No early {triggering}}}

\begin{lemma}[Decreasing maximum]
\begin{equation}
\max _{k}\Upsilon_i(k,t) \leq \max _{k}\Upsilon_i(k,t-1),
\label{DecreasingMaximum}
\end{equation}
\label{lemma:max}
\end{lemma}
\vskip -0.25in
where $\Upsilon_{i}(k,t):=\Pr (Y_{i,\theta}\left( t \right) = k)$.
This property can easily be proven by inserting the definition of $p_{i}$ into the following recursive formula derived from both the law of total probability and the definition of the Poisson-binomial distribution 

\begin{property}[Recursion on $k$, $t$]~\\
\begingroup\makeatletter\def\f@size{8}\check@mathfonts
\def\maketag@@@#1{\hbox{\m@th\large\normalfont#1}}%
\begin{equation}
\Upsilon_i(k,t) = \!
\begin{cases}
\left(1\!-\!p_{i}\! \left( t\right)\right)\! \Upsilon_{i}(k,t\!-\!1) \! & \!  k\!=\!0 \! \\ 
\left(1\!-\!p_{i}\! \left( t\right)\right)\! \Upsilon_{i}(k,t\!-\!1) + p_{i}\! \left( t\right)\!\Upsilon_{i}(k\!-\!1,t\!-\!1)  \! & \! k\!>\!0\! \\ 
\end{cases}
\label{RecursiveFormula}
\end{equation}
\endgroup
where $\Upsilon_{i}(k,0) = \mathds{1}_{k=0}$.  
\end{property}

Lemma~\ref{lemma:max} reveals that once the mass of $Y$ is dispersed, it cannot be reconcentrated.
Indeed, the variance of $\Upsilon_i(k,t)$ is nondecreasing with $t$:
\begin{equation}
\sigma^2_{\Upsilon_i(\cdot,t)} -\sigma^2_{\Upsilon_i(\cdot,t+1)} = (1-p_{i}(t+1))p_{i}(t+1)\ge 0.
\end{equation}

This second constraint clearly sets our approach apart from standard recurrent models, which can freely update their mass distribution over time.

\begin{lemma}[First upper bound]
\begin{equation}
\max _{k}\Upsilon_i(k,t) \leq \tfrac{1}{2} + \min_{j \leq t} \| \tfrac{1}{2} - p_{i}\!\left( j\right)\|.
\label{FirstBound}
\end{equation}
\end{lemma}

Indeed, as the ordering of the independent Bernoulli distributions up to time $t$ has no impact on the final distribution~$\Upsilon_i(k,t)$, the $p_{i}(\cdot)$ satisfying $\max _{j \leq t} \| \tfrac{1}{2} - p_{i}(j) \|$ can be placed first. The lemma then follows from (\ref{DecreasingMaximum}) and (\ref{RecursiveFormula}).

This inequality indicates that even a single prediction $p_{i}(\cdot)$ around~$\tfrac{1}{2}$ can cause the maximum of $Y$ to drop permanently. This bound~(\ref{FirstBound}) is loose as it only derives from a single $p_{i}(\cdot)$; according to the decreasing maximum property (\ref{DecreasingMaximum}), the rest of the $p_{i}(\cdot)$ can only reinforce this effect.

\subsection{Consequence} These first results can appear abstract. However, the connection between distribution upper-bounds and detection performance becomes evident once the definition of the \brown{LoCo}-loss function is restated:
\begin{equation}
\begin{split}
L(\theta) &= - \sum_{i}  \log  \left( \Pr \left( Y_{i,\theta} = y_{i}\mid\mat{x}_{i} \right)\right) \\
&= - \sum_{i}  \log  \left(  \Upsilon_i(y_{i},T_{i}) \right)  \\
&\stackrel{\text{(\ref{FirstBound})}}{\geq} - \sum_{i}  \log  \left( \tfrac{1}{2} + \min _{j \leq t} \| \tfrac{1}{2} - p_{i}\!\left( j\right) \| \right).
\end{split}
\end{equation}
In other words, if a sequence cross-entropy of $-\!\log\left( \alpha \right)$ is reported, then no estimated event probability $p_{i}(\cdot)$ can satisfy $\alpha \!\leq\! \tfrac{1}{2}\!  -\! \min_{j} \| \tfrac{1}{2} - p_{i}(j) \|$. Thus, a more in-depth understanding of $\max _{k}\Upsilon_i(k,t)$ can help us uncover properties of the predicted event process as the learning progresses.

To this end, further upper-bounds could be derived using Petrov's theorem~\yrcite{petrov2007lower} for tail lower-bound of distributions with finite fourth moment. However, the complexity of the final statements overshadows its potential relevance. On the other hand, Le Cam's theorem~\yrcite{le1960approximation} combined with the properties derived so far yields a more interpretable result when applied to our problem.

\begin{property}[Sparse mass concentration]
The inequality derived below reveals that, as the loss decreases, small $p_{i}(\cdot)$ will quickly converge towards zero.
\begingroup\makeatletter\def\f@size{8}\check@mathfonts
\def\maketag@@@#1{\hbox{\m@th\large\normalfont#1}}%
\begin{equation}
\begin{split}
&\max _{k}\Upsilon_i(k,t) \stackrel{\text{(\ref{DecreasingMaximum})}}{\leq}  \min_{l\leq t} \max _{k}\Upsilon_i(k,l) \stackrel{\text{ind}}{=} \min_{\sigma, l\leq t}  \max _{k}\Upsilon_{i,\sigma}(k,l)
\\
&\stackrel{\text{Le Cam}}{\leq} \min_{\sigma, l\leq t} \max_{k} {\lambda_{i,\sigma,l}^k e^{-\lambda_{i,\sigma,l}} \over k!} + 2 \sum_{j=1}^l p_{i,\sigma}(j)^2 \\
& \stackrel{\text{def}}{=} \min_{\sigma, l\leq t} \max_{k} {\left[ \sum_{j=1}^l p_{i,\sigma}(j) \right]^k e^{-\left[ \sum_{j=1}^l p_{i,\sigma}(j) \right]} \over k!} + 2 \sum_{j=1}^l p_{i,\sigma}(j)^2,
\raisetag{53pt}
\end{split}
\label{LeCam}
\end{equation}
\end{property}
\endgroup
where $p_{i,\sigma}(\cdot)$ stands for $p_i(\cdot)$ after permutation $\sigma$. The same notation is used for $\Upsilon$ and $\lambda$. 
This property can be more easily interpreted if one considers the permutation $\sigma$ which sorts the $p_{i}(\cdot)$ in ascending order.
\begin{example*}
Let us suppose that the hundred smallest $p_i(\cdot)$ of a sequence are equal to $0.01$. Substituting into Equation~(\ref{LeCam}) then yields $\max _{k}\Upsilon_i(k,T_i) \leq \frac{1}{e} + 0.02$, which leads to a sizable cross-entropy value. Consequently, as the learning progresses, even the smallest $p_{i}(\cdot)$ have to decrease to avoid the $Y$ distribution to diffuse.
\end{example*}

\subsection{Convergence}

Assuming that the events are detectable and well-defined, precise and almost binary localization will emerge from the model constraints. First, the monotonically increasing behavior of the sequence $\left({Y}_{i,\theta}\left(t\right)\right)_{t=1}^{T_{i}}$ implies that all mass shifts are permanent, thus preventing the model from triggering early. Second, Le Cam's inequality~(\ref{LeCam}) implies a weak mass concentration property. Indeed, unlike most benchmark models, a detection cannot be split into numerous small $p_{i}(\cdot)$ contributions. Most of the mass for a single event is thus concentrated within a few time steps. Third, these previous results combined with the first upper-bound~(\ref{FirstBound}) yields a \textit{strong mass concentration property}. Indeed, as contributions cannot be dispersed into small pieces and as predictions have values around the 0 and 1 extremes, a single $p_{i}(\cdot)$ will consequently contain most of the mass for a particular event as the loss decreases. \brown{\highlight{Strong mass concentration}}

\brown{Finally, there is no clear-cut theoretical explanation as to why late bias cannot occur, aside from the argument that the network would have to allocate large parts of its resources to keep triggers in memory. However, this issue can be cleanly addressed by feeding sequences of different lengths. Indeed, as the network never knows when the sequence will be ending, it cannot delay its decision; if it did, detections towards the end of the sample would be missed causing  loss surges. \highlight{No systematic late bias}}

In summary, if the model accurately learns to count occurrences and if the events are detectable, then a coherent localization will emerge naturally.

\section{Implementation Remarks}
\label{sec:sectionImplementation}

The indirect nature of the learning process presents some additional implementation challenges in comparison to more traditional models. Robust solutions to these issues such as loss computation or weight initialization will be discussed in this section.

\subsection{Loss Computation}

The Poisson-binomial distribution is the main component of the \brown{LoCo}-loss function; its efficient and accurate computation is thus crucial for the learning process. Evidently, the closed-form definition~(\ref{poissonBinomialDistribution}) can be computed without any difficulty on short-time horizons. However, the computational burden becomes unbearable for longer time frames due to the exponential nature of its \mbox{complexity}.

Numerous solutions have been developed to overcome this specific issue. First, approximation-based methods \cite{le1960approximation, roos2001binomial} are efficient, but can directly be discarded, since an exact computation of the loss is imperative for gradient descent learning.  Secondly, alternative closed-form formulas based on Fourier transforms are too complex for our application \cite{fernandez2010closed}. Finally, various recursive formulas have been derived \cite{howard1972discussion, shah1973distribution, gail1981likelihood, chen1994weighted, chen1997statistical}. 

After consideration and testing, the choice was set on the recurrence (\ref{RecursiveFormula}) discussed above.
Its numerical stability and simplicity outweighs its rather weak $O(T_i^2)$ complexity, while its convolutional form offers an elegant implementation solution. (See also~\cite{howard1972discussion, gail1981likelihood} for a more general case.)

\paragraph{Mass Thresholding} The extent of $Y_i$'s sample space is naturally bounded by $T_i$. However, imposing an even stricter bound $k_{\mathrm{max}}$ on the number of bins --- which amounts to truncating $Y$ from the right --- can be beneficial from a practical standpoint. Indeed, not only does the complexity drop from $O(T_{i}^2)$ to $O(k_{\mathrm{max}} T_{i})$, but the computation of the cross-entropy is also simplified by using the same fixed number of bins for each training sample. The only modification required involves the last bin, which must contain all the remaining mass above the threshold:
\vskip -0.15in
\begin{equation}
\widetilde{\Upsilon}_i(k_{\mathrm{max}},t) = \hspace{-1em}\sum_{\quad j\geq k_{\mathrm{max}}} \hspace{-1em}\Upsilon_i(j,t).
\end{equation}

\subsection{Weight Initialization}

The multiplicative nature of the Poisson-binomial distribution calls for caution when initializing the network's weights. Indeed, it is essential to avoid extreme $p_{i}$ values which may cause cross-entropy surges and exploding gradients. Traditional weight initialization methods such as Xavier~\cite{glorot2010understanding}  or He~\yrcite{he2016deep} can nevertheless be used without  any concern by simply adding the following initial bias to the pre-sigmoid predictions:
\begin{equation}
\log \left( \frac{1 - \omega^{1/T_{i}}}{\omega^{1/T_{i}}} \right).
\end{equation}
Such initialization produces a balanced initial $Y_i$ distribution with approximately $\omega$ mass on the first bin ($\omega$ should be chosen to be substantially different from both $0$ and $1$).

\subsection{Overfitting Considerations}

Similarly to any deep learning-based approaches, the proposed model can be subject to overfitting when model complexity and dataset size are not properly balanced. However, in such a case, not only the out-of-sample count predictions, but also the accuracy of the predicted event process might deteriorate in both in- and out-of-sample settings. Hence, examination of the estimated in-sample event process can be used to fine-tune the model size.

\clearpage

We now demonstrate the effectiveness of our model in a series of \brown{challenging} experiments.

\section{Drum Transcription Experiments}
\label{DrumTrascription}

Drum transcription consists in detecting and classifying drum hits in audio extracts. In this section, a few standard experiments as proposed by Wu~et~al.~\yrcite{wu2018review} are conducted using our weakly-supervised model. Although their review only includes fully-supervised approaches, a comparison with their reported results can act as a relevant indicator of our model's effectiveness.

The interest for these experiments is two-fold. First, given the instantaneous and highly localized nature of drum hits, the model can be tested under minimal violation of our model assumptions. Secondly, the task~--- which requires predictions to be within $50\si{\milli\second}$ of ground truth~\cite{wu2018review}~--- challenges the temporal localization precision of our model.

\setlength\dashlinedash{0.2pt}
\setlength\dashlinegap{1.8pt}
\setlength\arrayrulewidth{0.2pt}

\begin{table*}[!b]

\caption{Drum Detection Results. Comparison between our \textit{weakly}-supervised model and \textit{fully}-supervised models evaluated in~\cite{wu2018review}. The F$_1$ scores per instrument (KD/SD/HH), as well as the average precision, recall, and overall F$_1$ are displayed, [$\%$]. For details: \textsc{RNN}, \textsc{ReLUts} \cite{vogl2016recurrent}, \textsc{RNN}, \textsc{tanhB} \cite{southall2016automatic}, \textsc{GRUts} \cite{vogl2017drum} and \textsc{lstmpB} \cite{southall2017automatic}.}
\label{DTD}
\vskip 0.15in
\begin{center}
\begin{small}
\begin{sc}
\begin{tabular}{cl}
\multicolumn{2}{c}{}\\
\toprule
&Method \\
\midrule
\multirow{6}{*}[-0.4ex]{\rotatebox{90}{Random}} & RNN \\
&tanhB \\
&ReLUts \\
&lstmpB \\
& GRUts \\[0.3ex]
& \SkyBlue{ours \brown{(LoCo)}} \\
\midrule

\multirow{6}{*}[-0.4ex]{\rotatebox{90}{Subset}} & RNN\\
&tanhB	 \\
&ReLUts	 \\
&lstmpB	 \\
&GRUts	  \\[0.3ex]
& \SkyBlue{ours \brown{(LoCo)}}  \\
\bottomrule
\end{tabular}
\rule{0.2em}{0ex}
\begin{tabular}{ccc:cc:c}
\multicolumn{6}{c}{D-DTD dataset}\\
\toprule
KD & SD & HH & Pre & Rec & F$_1$ \\
\midrule
97.2 & 92.9 & 97.3 & 95.7 & 96.9 & 95.8\\
95.4 & 93.1 & 97.3 & 93.9 & 97.1 & 95.3\\
86.6 & 93.9 & \textbf{97.7} & 92.7 & 95.0 & 92.7\\
\textbf{98.4} & \textbf{96.7} & 97.4 & \textbf{97.7} & \textbf{97.6} & \textbf{97.5} \\
91.4 & 93.2 & 96.2 & 91.8 & 97.2 & 93.6\\[0.3ex]
\SkyBlue{96.0} &  \SkyBlue{90.4} & \SkyBlue{97.1} & \SkyBlue{95.1} & \SkyBlue{93.9} & \SkyBlue{94.5} \\
\midrule
88.0	& 85.3	& 93.2	& 86.0  & 95.1	& 88.9 \\
 91.9	& 89.9	& \textbf{94.4}	& \textbf{95.1}	& 91.2	& 92.1 \\
 91.2	& \textbf{90.9}	& 91.6	& 89.2	& \textbf{95.8}	& 91.2 \\
 \textbf{96.0}	& 88.7	& 93.8	& 93.8	& 94.0  & \textbf{92.8} \\
89.1	&   90.6	& 91.7	& 89.6	& 94.2	& 90.5 \\[0.3ex]
 \SkyBlue{88.0} & \SkyBlue{79.5} & \SkyBlue{93.9} & \SkyBlue{90.6} & \SkyBlue{84.3} & \SkyBlue{87.1} \\
\bottomrule
\end{tabular} 
\rule{0.4em}{0ex}
\begin{tabular}{ccc:cc:c}
\multicolumn{6}{c}{D-DTP dataset}\\
\toprule
 KD & SD & HH & Pre & Rec & F$_1$ \\
\midrule
 \textbf{94.7}	& 79.5	& 88.3	& 84.1	& \textbf{93.3}	& 87.5\\
92.4	& 84.6	& 87.1	& 86.3	& 92.1	& 88.0\\
91.3	& 83.8	& 85.2	& 83.7	& 92.3	& 86.8\\
94.4	& 84.1	& 91.4	& 90.8	& 90.8	& \textbf{90.0}\\
94.2	& \textbf{87.1}	& 87.7	& 88.6	& 92.7	& 89.7\\[0.3ex]
\SkyBlue{92.3} & \SkyBlue{81.2} & \textbf{\SkyBlue{93.0}} & \textbf{\SkyBlue{90.9}} & \SkyBlue{87.1} & \SkyBlue{88.9} \\

\midrule
 \textbf{91.0}	& 57.8	& 82.2	& 72.8	& \textbf{88.3}	& 77.0\\
82.7	& 61.6	& 84.8	& 74.1	& 83.8	& 76.4\\
79.4	& 62.1	& 80.8	& 69.6	& 84.2	& 74.1\\
85.8	& \textbf{68.8}	& 83.7	& 78.3	& 84.7	& \textbf{79.4} \\
87.7	& 62.3	& 79.4	& 73.0	& 85.2	& 76.5\\[0.3ex]

\SkyBlue{84.9} & \SkyBlue{59.4} & \textbf{\SkyBlue{90.0}}  & \textbf{\SkyBlue{84.8}} & \SkyBlue{73.5} & \SkyBlue{78.1}  \\
\bottomrule
\end{tabular}
\end{sc}
\end{small}
\end{center}
\vskip -0.1in
\end{table*}

\subsection{Experiment Specifications}
\label{subsection:drumspecification}

\paragraph{Dataset}  The model is evaluated on two different datasets: IDMT-SMT-Drums~\cite{dittmarreal} and ENST Drums~\cite{gillet2006enst}. The latter is considered more challenging than the former as it includes a wider variety of simultaneously playing drums \brown{(i.e. more background clutter)}.

As the total number of tracks is limited, each audio extract is first split into $1.5\si{\second}$ segments to artificially increase the dataset size. For each of these snippets, the total number of occurrences for each drum type (hi-hat (HH), snare drum (SD), and bass kick drum (KD)) are then determined and used as training labels, thus discarding any localization information. \brown{($T_{\max}$: 400, $k_{\max}$: 31)}

\paragraph{Architecture} The network architecture is kept simple as the datasets are quite limited in size. First, the representation learning part of the network is composed of six ($3\times 4$) convolutional layers with 8 to 16 filters intertwined with max-pooling layers and ReLU activations. Secondly, the recurring unit is comprised of a 24-unit LSTM which is then directly followed by a final 16-node fully-connected prediction layer. 
Finally, as an additional effort to simplify the learning process, three different models are trained to detect each drum type separately.

\paragraph{Training} Mel-spectrograms~\cite{stevens1937scale} stacked together with their first derivatives are used as model input. In addition, data augmentation in the form of sample rate variations is applied during both training and inference.  The \brown{LoCo}-loss described in Section~\ref{Loss} is optimized using the Adam \mbox{algorithm \cite{kingma2014adam}}.

\paragraph{Evaluation} The \textsl{Eval Random} and \textsl{Eval Subset} of the D-DTD and D-DTP tasks as defined by Wu~et~al.~\yrcite{wu2018review} are selected for the evaluation~--- detailed information on the exact protocol can be found in their work. The \textsl{Eval Random} task assesses the model performance on similar data, whereas the \textsl{Eval Subset} tests their generalization capability. Cross-validated results are obtained by aggregation of six independent runs. 

The full implementation and additional details can be found on the paper's website\footnote{\label{noteWeb}\brown{http://users.cs.cf.ac.uk/SchroeterJ1/publications/LoCo}}.

\subsection{Results}

As shown in Table~\ref{DTD}, the proposed model is competitive against fully-supervised state-of-the-art drum transcription methods~\cite{wu2018review} in most of the experiments. Our weakly-supervised method achieves precise localization \textit{without any localization prior}.  

Remarkably, the localization error is often much smaller than the $50\si{\milli\second}$ tolerance. For instance, the mean F$_1$-score of hi-hats on the D-DTD \textsl{Eval Random} task drops only from  $97.1\%$ to $96.3\%$ when the tolerance is reduced to $20\si{\milli\second}$. In this setting, an impressively tight localization is achieved as demonstrated by a standard deviation of only $4.35\si{\milli\second}$ for the distance between true and predicted hits.

The proposed model reaches high levels of precision as is especially apparent from the results on the more challenging D-DTP dataset. This confirms that as long as the number of occurrences is estimated correctly, precise localization emerges naturally.
As a design choice, a fairly narrow network was preferred over a larger architecture due to its inherent robustness, which explains the slightly lower recall.
If required, the imbalance between precision and recall can however be alleviated by performing model ensembling with a low selection threshold. 

Overall, the model displays outstanding performance as it achieves results comparable to those of fully-supervised methods while only using occurrence counts as training labels (without any localization information).

\section{Piano Onset Detection Experiment}
\label{sectionPiano}

Note onset detection is an essential part of music transcription. 
However, with 88 different channels and complex interactions, the specific task of piano onset detection is particularly challenging. In this section, the experiment conducted by Hawthorne~et~al.~\yrcite{hawthorne2017onsets} based on the MAPS database \cite{emiya2010multipitch} is replicated using our weakly-supervised approach. Even though their model also predicts offsets and note velocities, only onset times are considered for this experiment.

\subsection{Experiment Specifications}

\paragraph{Dataset} The MAPS database is used for this evaluation. As in \cite{hawthorne2017onsets}, the synthesized pieces are used for training, whereas the Disklavier pieces are used for testing. In addition, samples containing only single notes and chords are also discarded producing a more challenging and more realistic training set. 

As for the drum experiment, each audio extract is split into $1.5\si{\second}$ segments to artificially increase the dataset size and only occurrence counts are used for training. 

\paragraph{Architecture and Training} The model architecture is similar to the one used in Section~\ref{subsection:drumspecification} for drum transcription. The only difference resides in the number of convolutional filters and recurring units, which is increased by a factor between 2 and 4. Separate models are trained to each detect a different band of 10 consecutive \brown{pitches}. Finally, data augmentation is implemented in the form of time stretching and extract stacking (i.e., playing two samples simultaneously). \brown{\mbox{($T_{\max}$: 400, $k_{\max}$: 42)}}

\subsection{Results}

As outlined in Table~\ref{results:piano}, our model not only achieves onset localization performance close to that of the fully-supervised state-of-the-art \cite{hawthorne2017onsets}, but also clearly outperforms the other tested approaches \cite{sigtia2016end, kelz2016potential} despite much weaker training labels. Once again, these results demonstrate the effectiveness of our weakly-supervised approach, which yields precise localization (within $50\si{\milli\second}$) without any localization prior.

\begin{table}[h!]
\setlength{\tabcolsep}{5.6pt}
\caption{Piano Onset Detection Results. Comparison between our \textit{weakly}-supervised approach and \textit{fully}-supervised models evaluated in~\cite{hawthorne2017onsets} on the MAPS dataset. For consistency, final metrics are computed as the mean over all pieces' score. [$\%$].}
\label{results:piano}
\vskip 0.0in
\begin{center}
\begin{small}
\begin{sc}
\begin{tabular}{lcc:c}
\toprule
Method & Pre & Rec & F$_1$ \\
\midrule
Sigtia et al.\yrcite{sigtia2016end} & \tbcolor{44.97}	& \tbcolor{49.55}	& \tbcolor{46.58} \\
Kelz et al.\yrcite{kelz2016potential} & \tbcolor{44.27}	& \tbcolor{61.29}	& \tbcolor{50.94} \\
Hawthorne et al.\yrcite{hawthorne2017onsets} & \tbcolor{\textbf{84.24}}	& \tbcolor{\textbf{80.67}}	& \tbcolor{\textbf{82.29}}\\[0.3ex]
\SkyBlue{ours \brown{(LoCo)}} & 	\SkyBlue{76.22} & \SkyBlue{68.61}	& \SkyBlue{71.99} \\

\bottomrule
\end{tabular}
\end{sc}
\end{small}
\end{center}
\vskip -0.25in
\end{table}

A more in-depth analysis reveals that the model displays excellent results on medium and high notes, while being slightly less effective in coping with the small occurrence rate and the more complex harmonic structures of lower notes. The application of specific spectral transformations or an artificial increase in the number of lower notes in the dataset would certainly alleviate this effect and make our method even more competitive. (However, this is beyond the scope of this paper.)

\section{Digit Detection Experiment}
\label{sectionDigit}

In this section, an application of our model to object detection in images is presented in order to assess both representation learning and localization learning separately.

\subsection{Approach}

In order to fulfill our model's input requirements, the original image ($\real^{W \times H \times d}$ space) is sampled by taking windows of size $w\times h$ along a space-filling curve~\cite{peano1890courbe}, thus transforming
the image into a sequence of sub-images 
($\real^{T\times(w \times h \times d)}$ space). Specifically, the Hilbert curve~\yrcite{hilbert1935stetige} is used for this experiment. In this setting, the recurrent unit has the challenging task of simultaneously learning space mapping, detection, and recognition.

\subsection{Experiment Specifications}

\begin{figure}[b!]
\vskip -0.12in
\begin{center}
\centerline{\includegraphics[width=7cm]{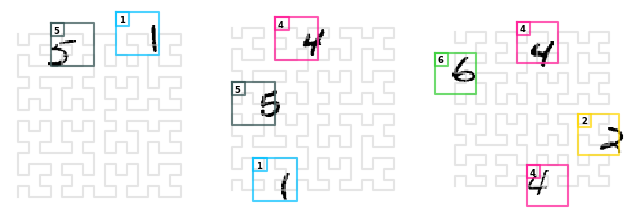}}
\caption{Out-of-sample predicted MNIST digit detection. \brown{(Raw prediction without postprossesing nor non-maximum suppression.)}}
\label{digit_localization}
\end{center}
\vskip -0.20in
\end{figure}

{\bf Dataset} The well-known MNIST~\cite{lecun1998gradient} dataset is used to generate samples for this experiment. More specifically, each synthetic image is comprised of non-overlapping digits sampled from MNIST and placed uniformly at random as illustrated in Figure~\ref{digit_localization}. The original train-test split is kept. Once again, only the number of occurrences of each digit is provided for training.

{\bf Network Architecture} The representation learning part is identical to the convolutional layers of the VGG-13 architecture~\cite{simonyan2014very}, while the localization part consists of a 48-unit LSTM. This is followed by a 24-node fully-connected prediction layer.

\subsection{Results}

The results of digit recognition, representation learning and localization learning are addressed separately.

\textbf{Digit Recognition} After the weakly-supervised training, our model achieves a $99.12\%$ single digit recognition accuracy, which is noticeably better than the fully-supervised VGG-13 score of $98.51\%$ even though both networks share an identical representation learning architecture. This result demonstrates that indirect localization learning is achieved without detriment to representation learning or recognition accuracy.

\begin{figure}[t]
\vskip 0.20in
\begin{center}
\centerline{\includegraphics[width=7.5cm]{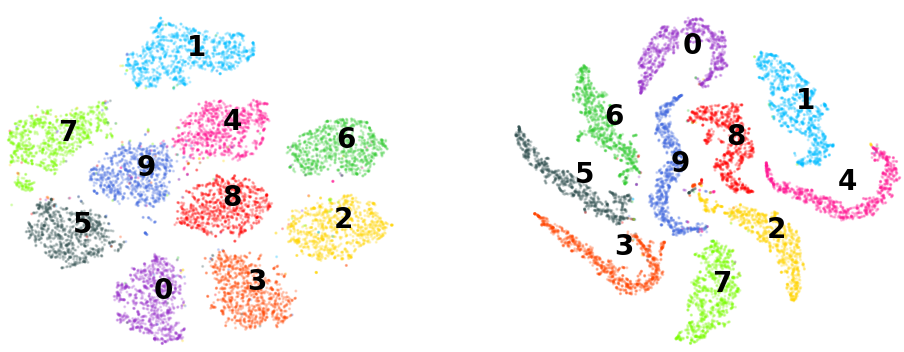}}
\caption{Digit Representations. Comparison of t-SNE digit feature representations resulting from the \textit{fully}-supervised VGG-13 architecture (left) and from our \textit{weakly}-supervised approach (right).}
\label{tsne_digits}
\end{center}
\vskip -0.50in
\end{figure}

\textbf{Representation Learning} In order to assess the representation learning alone, original ($28\times 28$) digit images are first fed to our network. The representations of the last convolutional layer are then selected and visualized using a t-SNE representation~\cite{maaten2008visualizing}. The same experiment is also conducted for the fully-supervised VGG-13 network. The comparative result can be observed in Figure~\ref{tsne_digits}. Overall, topological considerations aside, the discriminative nature of the representation is comparable for both approaches. 
In this case, the indirect nature of the learning process and the weaker training annotations do not affect the quality of the representations which are almost comparable to that of fully-supervised approaches.

\textbf{Localization Learning} The localization performance is evaluated by computing the mean absolute distance between true and estimated bounding box centers. This experiment results in a value of $9.04$ pixels, which is close to the granularity of the space-filling curve ($8$ pixels), demonstrating once again the effectiveness of our model.

{\bf Conclusion} Our model can learn both representation and object detection simultaneously in an indirect fashion. In this case, the performance gap for using a weakly-supervised rather than a fully-supervised approach is minimal despite much weaker annotation requirements.

\subsection{Limitations}

The detection scale is defined by the size of the sub-images. However, this limitation can be lifted by extending the proposed approach to a multi-scale one or by replacing the fixed space-filling curve by a learnable adaptive-scale scanning process using reinforcement learning \cite{mnih2015human}.

\section{Conclusion}

In this work, we have shown how implicit model constraints can be used to ensure that accurate localization emerges as a byproduct of learning to count occurrences.
Experimental validation of the model demonstrates its competitiveness against fully-supervised methods on challenging tasks, despite much weaker training requirements.
In particular, both precision in the order of a few milliseconds in the drum detection task and strong performance in the piano transcription experiment have been achieved without any localization prior.
Furthermore, the proposed approach has displayed the ability to naturally learn meaningful representations while learning to count.

The properties of the model can be leveraged for further applications. For instance, the precise localization power of the model can be exploited for enriching any sequential data whenever it contains imprecise or poorly defined localization information, while the mass concentration property can act as a regularizer in other models to ensure sharp mass converge towards well-localized points.

\section*{Acknowledgement}

\brown{
We gratefully acknowledge the support of NVIDIA Corporation with the donation of the GPU used for this research.}

\bibliography{references}
\bibliographystyle{icml2019}

\end{document}